\documentclass[letterpaper, 10 pt, conference]{ieeeconf}  

\IEEEoverridecommandlockouts                              

\usepackage{graphics} 
\usepackage{epsfig} 
\usepackage{times} 
\usepackage{amsmath} 
\usepackage{amssymb}  
\usepackage{xcolor}
\usepackage{booktabs}
\usepackage{multirow}
\usepackage{cite}
\usepackage{subcaption}
\usepackage{makecell}
\usepackage{algorithm,algorithmic}
\let\labelindent\relax
\usepackage{enumitem}
\usepackage{url}
\usepackage{hyperref}

\newcommand{\trans}{\mathsf{T}}
\newcommand{\smooth}{\text{smooth}}
\DeclareMathOperator{\tr}{tr}

\title{\LARGE \bf
ComFree-Sim: A GPU-Parallelized Analytical Contact Physics Engine for Scalable Contact-Rich Robotics Simulation and Control
\vspace{-5pt}
}

\makeatletter
\g@addto@macro\@maketitle{
\setcounter{figure}{0}
  \vspace{-2pt}
  \begin{figure}[H]
  \setlength{\linewidth}{\textwidth}
  \setlength{\hsize}{\textwidth}
  \centering
  \includegraphics[width=\textwidth]{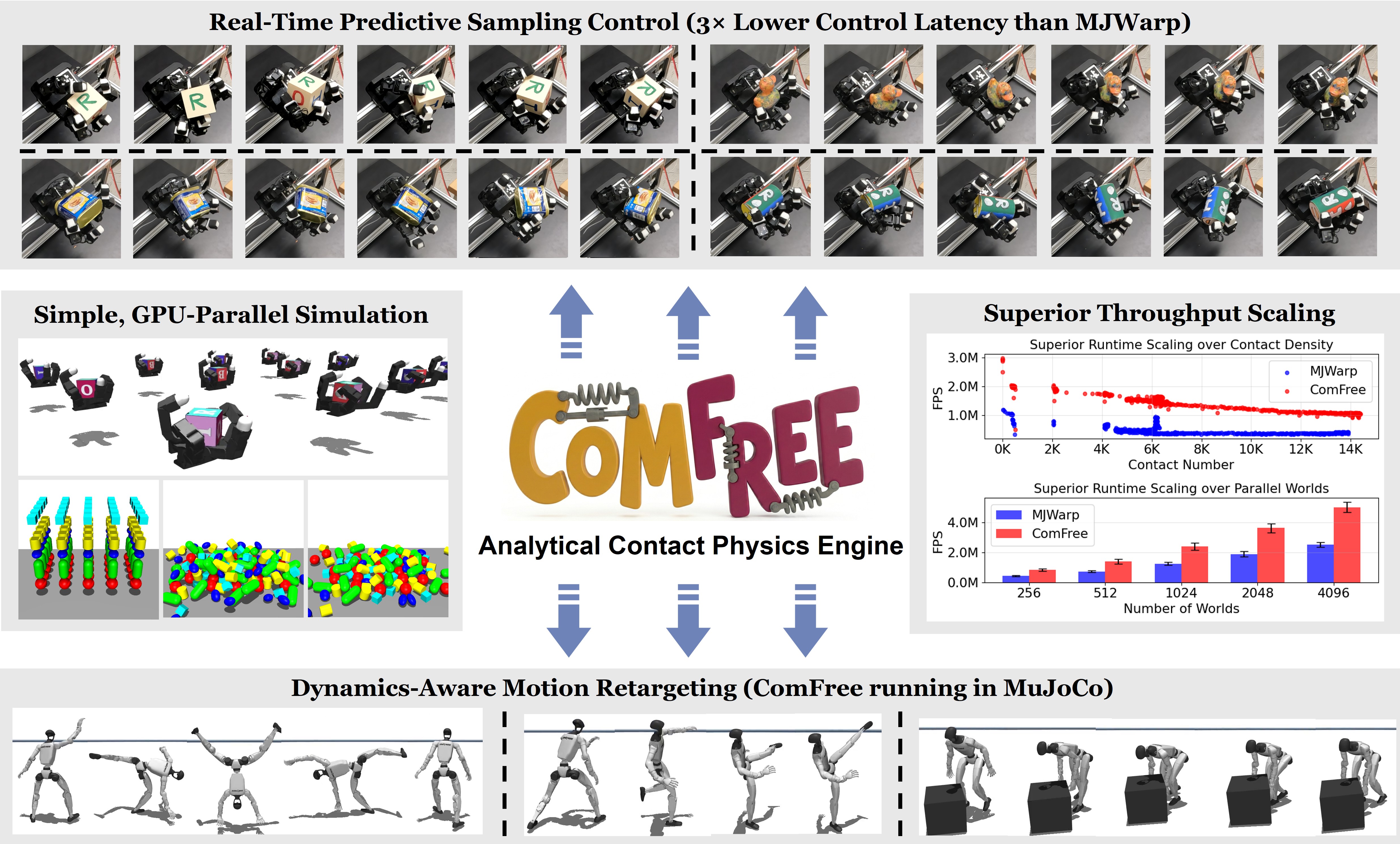}
  \vspace{-15pt}
  \caption{Performance overview of the  ComFree-Sim. In the second row, it shows 2--3$\times$ higher throughput than MuJoCo Warp (MJWarp) in dense contact  simulation; in the first row, it enables low-latency model predictive control for  in-hand dexterous manipulation; and in the third row, it is used for dynamics-aware motion retargeting with MuJoCo as rollout environments. 
  }
  \label{teaser_fig}
  \end{figure}
  \vspace{-10pt}
}

\author{Chetan Borse$^{*1}$ \quad Zhixian Xie$^{*1}$ \quad Wei-Cheng Huang$^{2}$
 \quad Wanxin Jin $^{\dagger1}$
 \\[5pt]
 $^{1}$ Intelligent Robotics and Interactive Systems (IRIS) Lab,
Arizona State University \\
 $^{2}$ Siebel School of Computing and Data Science, University of Illinois at Urbana-Champaign \\
 $^{*}$Co-first author with equal contribution,\quad $^{\dagger}$ Corresponding author (wjin@asu.edu)\\
  Project  website: \url{https://irislab.tech/comfree-sim/}.
\thanks{C.B. developed the initial  implementation of ComFree-Sim and hardware MPC. Z.X. conducted simulation, MPC, and retargeting experiments, and writing. W.C.H. contributed to MPC, discussions, and writing. W.J. led the project and contributed to coding ComFree-Sim.}
}

\begin{document}

\maketitle
\thispagestyle{empty}
\pagestyle{empty}

\begin{abstract}
Physics simulation for contact-rich robotics is often bottlenecked by contact resolution: mainstream engines enforce non-penetration and Coulomb friction via complementarity constraints or constrained optimization, requiring per-step iterative solves whose cost grows superlinearly with contact density. We present ComFree-Sim, a GPU-parallelized analytical contact physics engine built on complementarity-free contact modeling. ComFree-Sim computes contact impulses in closed form via an impedance-style prediction--correction update in the dual cone of Coulomb friction. Contact computation decouples across contact pairs and becomes separable across cone facets, mapping naturally to GPU kernels and yielding near-linear runtime scaling with the number of contacts. We further extend the formulation to a unified 6D contact model capturing tangential, torsional, and rolling friction, and introduce a practical dual-cone impedance heuristic. ComFree-Sim is implemented in Warp and exposed through a MuJoCo-compatible interface as a drop-in backend alternative to MuJoCo Warp (MJWarp). Experiments benchmark penetration, friction behaviors, stability, and simulation runtime scaling against MJWarp, demonstrating near-linear scaling and 2--3$\times$ higher throughput in dense contact scenes with comparable physical fidelity. We deploy ComFree-Sim in real-time MPC for in-hand dexterous manipulation on a real-world multi-fingered LEAP hand and in dynamics-aware motion retargeting, demonstrating that low-latency simulation yields higher closed-loop success rates and enables practical high-frequency control in contact-rich tasks. 
\end{abstract}

\section{INTRODUCTION}

Scalable physics simulation has become core infrastructure for modern robotics, enabling large-scale data generation for policy training, parallelized sampling for model predictive control (MPC), and end-to-end differentiable pipelines for system design. Recent GPU-accelerated simulators and system-level platforms have greatly improved parallel simulation throughput, lowering the barrier to training and evaluating increasingly complex tasks. Yet, for \emph{contact-rich} manipulation and locomotion, simulation remains bottlenecked by the same fundamental difficulty: resolving intermittent, unilateral, frictional contacts accurately and efficiently. 

A rigid-body simulation step typically includes collision detection, contact resolution, and time integration \cite{Bender2012interactive,le2024contactmodels}. Among these, \emph{contact resolution}, which computes contact impulses and   post-contact velocities, most strongly impacts physical fidelity, numerical stability, and runtime \cite{horak2019similarities}. Main approaches for contact resolution  enforce non-penetration and Coulomb friction via the complementarity constraints or equivalent constrained optimization \cite{stewart1996implicit,anitescu2006optimization,todorov2012mujoco}, resulting in iterative per-step solutions whose cost often grows superlinearly with  contact counts. This scaling is prohibitive for batched, differentiable simulation and online planning. Even state-of-the-art GPU-parallelized phsyics engines, e.g., MJWarp \cite{mujoco_warp} can exhibit substantial step-time growth in dense contact scenes, complicating real-time deployment.

To overcome the iterative-solver contact resolution bottleneck, this paper presents \textbf{ComFree-Sim}, a GPU-parallelized analytical contact backend for contact-rich simulation. ComFree-Sim builds on \emph{complementarity-free} contact modeling \cite{jin2024complementarity}, which resolves contact impulses in closed form through a prediction-based impedance mechanism in the \emph{dual cone} of Coulomb friction. A key property of this formulation is \emph{decoupling}: contact resolution is independent across contact pairs and separable across cone facets. This structure maps directly to GPU parallelism, reducing computation complexity from superlinear to linear with respect to contact count. Our empirical results demonstrate that this lightweight analytical formulation preserves physical realism without noticeable degradation. By eliminating repetitive optimization routines, it avoids solver-induced artifacts while preserving frictional feasibility by construction. Importantly, ComFree-Sim is implemented in Warp and exposed through a MuJoCo-compatible interface, enabling it to serve as an accelerated drop-in replacement backend for MJWarp. This paper makes three main contributions:

\begin{itemize}[leftmargin=10pt]
\item \textbf{Unified  analytical 6D frictional contact modeling.} We extend the complementarity-free point contact model to a  6D  formulation that captures tangential, torsional, and rolling friction within an analytical dual-cone contact resolution framework.

\item \textbf{GPU parallelization with stable impedance heuristics.} We develop  GPU implementation for the analytical contact resolution and introduce an effective dual-cone impedance heuristic. This preserves a lightweight user-facing parameterization while also supporting learning-based dynamic impedance adaptation \cite{wang2026gswm}.

\item \textbf{Evaluation from simulation to hardware.} We benchmark ComFree-Sim against MJWarp on penetration depth, various friction behaviors, numerical stability, and  runtime scaling. We demonstrate linear scaling and 2--3 times higher throughput in contact-rich scenes with comparable physical fidelity. We deploy ComFree-Sim for real-time MPPI-based MPC for dexterous in-hand manipulation on multi-fingered LEAP hand hardware, showing that faster rollouts translate directly into higher closed-loop success rates and improved practical deployability.
\end{itemize}  

Together, these results suggest that analytical, lightweight contact resolution can provide a highly competitive and scalable alternative to iterative complementarity-based backends, particularly for the high-frequency, contact-rich rollouts that underpin modern learning and control.

\section{Background and Related Work}
\subsection{Brief Background of Rigid Body Simulation}

Mainstream rigid-body simulators for robotics typically follow a time-stepping pipeline with three stages \cite{Bender2012interactive,le2024contactmodels}: \textbf{collision detection}, \textbf{contact resolution}, and \textbf{time integration}. Collision detection identifies potentially interacting body pairs and computes geometric contact quantities (e.g., contact points, normals, signed distances), often using a broad phase (BVH or spatial hashing \cite{wald2007ray}) followed by a narrow phase e.g., GJK \cite{gilbert1988fast}, EPA \cite{van2001proximity}, or SDF methods \cite{macklin2020local,yang2025contactsdf}. Contact resolution then computes contact impulses and post-contact velocities that satisfy force--motion constraints (Newton’s laws and frictional contact mechanics); these are commonly posed as complementarity-style constraints \cite{todorov2012mujoco,stewart1996implicit,anitescu2006optimization} to capture the unilateral, hybrid nature of frictional contact, and are typically solved \emph{iteratively} in practice. Finally, time integration advances the state using the resolved post-contact velocities; for broader comparisons, see recent surveys \cite{le2024contactmodels,horak2019similarities}.

While this pipeline is shared by most simulators, the dominant differences in physical realism, numerical stability, and runtime often arise in \emph{contact resolution} \cite{horak2019similarities}, where both the contact formulation \cite{todorov2012mujoco,stewart1996implicit,anitescu2006optimization,castro2022unconstrained} and the solver can vary substantially. In this paper, we focus on an \emph{analytical} contact resolution method: by avoiding per-step complementarity constraints and heavy iterative solves, our approach computes contact impulses in closed form.

\subsection{Robotics Simulators and Platforms}
Early physics engines such as ODE \cite{smithode2008}, Bullet \cite{coumans2016pybullet}, and PhysX \cite{macklin2019small} were developed mainly for animation and games, prioritizing visually plausible motion over high-fidelity contact dynamics \cite{erez2015simulation};  their \emph{contact resolution} often adopts simplified contact models and complementarity-style approximations. In contrast, robotics-oriented simulators such as MuJoCo \cite{todorov2012mujoco}, Pinocchio \cite{carpentier2019pinocchio}, and Drake \cite{drake} emphasize efficiency and physics fidelity for planning, control, and policy learning. In particular, MuJoCo and Drake enforce full friction-cone constraints and cast contact resolution as a cone complementarity problem solvable via convex optimization \cite{anitescu2006optimization}. Driven by large-scale training and hardware acceleration, these simulators have been further ported to GPUs/TPUs, with MJX and MJWarp built on MuJoCo and Isaac Gym/Sim \cite{makoviychuk2021isaac, NVIDIA_Isaac_Sim} built on PhysX as representative examples.

In parallel, modular system-level platforms provide out-of-the-box tasks and benchmarks for robot learning and evaluation. Rather than introducing new contact models/solvers, they typically {build on existing physics backends} and focus on curated assets, task definitions, and integrated modules (e.g., rendering, data generation). Examples include LIBERO \cite{liu2023libero},  OmniGibson \cite{li2023behavior},  ManiSkill \cite{Tao2025maniskill3}, IsaacLab \cite{isaaclab2025}, and more recent Mjlab \cite{zakka2026mjlab}.  These systems are popular because they support diverse robots and large-scale GPU parallel  backends such as PhysX, MJX, and MJWarp,   scaling policy learning  \cite{andrychowicz2020learning,kaufmann2023champion,lee2020learning}. \emph{In contrast, rather than pursuing system-integration innovations, we focus on a new analytical contact-computation backend with a GPU-parallelized implementation, designed to be  readily integrated into these task-centric platforms.}

\subsection{Why A Lightweight Yet Competitive Contact Backend?}


 A major computational bottleneck for scalable simulation is \emph{contact resolution}. Many simulators enforce contact force--motion constraints by repeatedly solving complementarity-style systems or equivalent convex programs, leading to \emph{superlinear} per-step cost that is often at least quadratic in the number of contacts (e.g., dense factorizations, projections, or repeated iterations). This becomes prohibitive for batched,  differentiable simulation, and online planning. Although recent parallel simulators enable online sampling-based predictive control with full-order simulation \cite{howell2022, li2025judo}, real-time results are still largely limited to contact-sparse or low-frequency control settings \cite{howell2022, li2025judo,li2025drop,Alvarez2025realtime}; extending to \emph{dense contact} and \emph{high-frequency} control remains challenging due to heavy contact-resolution solves.

The above demand motivates a \emph{lightweight} physics backend that avoids heavy per-step complementarity/optimization  and has the computational complexity scales \emph{linearly} with the number of contacts. Meantime, such a backend must remain \emph{competitive} with mature and widely used simulators (e.g., MuJoCo and their accelerated variants) in  stability and task performance, so it can serve as a practical drop-in engine for large-scale training and real-time control.

\section{ComFree-Sim: Analytical Contact Physics Computation and GPU Implementation}

Complementarity-free (ComFree) contact modeling was proposed in~\cite{jin2024complementarity} and subsequently demonstrated in contact-rich control and  learning~\cite{xie2026touch,wang2026gswm,xiepalm}. Our ComFree-Sim will extend it with (1) a unified 6D contact model covering tangential, torsional, and rolling friction, (2)  effective dual-cone impedance heuristics, and (3)  GPU parallel implementation \cite{macklin2022warp} as a drop-in backend alternative to MJWarp \cite{mujoco_warp}.

\subsection{Generic Complementarity-Free Contact Modeling}


Consider  multi-joint robot systems (robots, objects, environments) with dry joint friction, joint limits, and frictional or frictionless contacts. Over a timestep $dt$, the dynamics is
\begin{equation}
\label{equ:sys_dyn}
\boldsymbol{M}(\boldsymbol{q})\,d\boldsymbol{v} + \boldsymbol{c}(\boldsymbol{q},\boldsymbol{v})\,dt
= \boldsymbol{\tau}\,dt + \mathbf{J}(\boldsymbol{q})^{\top}\boldsymbol{\lambda}\,dt .
\end{equation}
Here, $\boldsymbol{q}$ and $\boldsymbol{v}$ are joint position and velocity; $\boldsymbol{M}(\boldsymbol{q})$ is the joint-space inertia; $d\boldsymbol{v}$ is the velocity increment; $\boldsymbol{c}(\boldsymbol{q},\boldsymbol{v})$ collects bias forces (Coriolis/centrifugal/gravity); $\boldsymbol{\tau}$ is the applied generalized force; $\mathbf{J}(\boldsymbol{q})$ is the contact Jacobian; and $\boldsymbol{\lambda}$ is the contact force/torque (wrench). We omit dependence on $(\boldsymbol{q},\boldsymbol{v})$ for $\boldsymbol{M}, \boldsymbol{c}, \mathbf{J}$  when clear. Since hard-contact Coulomb friction is ill-defined in continuous time (Painlev\'e's paradox), we interpret $\boldsymbol{\lambda}$ as the \emph{step-averaged} contact wrench, so $\boldsymbol{\lambda}\,dt$ is the contact impulse.

Complementarity-free contact modeling \cite{jin2024complementarity} computes the contact forces \emph{analytically}, avoiding per-step complementarity constraints and iterative optimization while automatically satisfying frictional constraints. Intuitively, it follows a \emph{prediction--correction} principle on the dual cone of the  friction cone: it first predicts the momentum/velocity under non-contact (smooth) forces, then applies a correction that resolves constraint violations (non-penetration and frictional feasibility) via a  residual-based impedance mechanism. For notation simplicity, we below present the single-contact case; the formulation naturally extends to multiple contacts.

Specifically, the one-step prediction of the system velocity only under non-contact (smooth) forces $\boldsymbol{\tau}$ and $\boldsymbol{c}$   is  
\vspace{-2pt}
\begin{equation}
\vspace{-2pt}
\label{equ.smooth_vel}
\boldsymbol{v}_{\smooth}^+ := \boldsymbol{v} + \boldsymbol{M}^{-1}(\boldsymbol{\tau}-\boldsymbol{c})\,dt.
\end{equation}

To handle various frictional behaviors in a unified manner, we consider (1) tangential (sliding) friction, (2) torsional friction about the contact normal, and (3) rolling friction in the tangential plane. We denote the normal force by $\lambda^{n}\in\mathbb{R}$, the tangential friction impulse by $\boldsymbol{\lambda}^{\mathrm{t}}\in\mathbb{R}^{2}$, the torsional friction moment by $m^{\mathrm{tor}}\in\mathbb{R}$, and  rolling friction moment by $\boldsymbol{m}^{\mathrm{roll}}\in\mathbb{R}^{2}$. The (primal) friction cone constraints are
\begin{equation}
\|\boldsymbol{\lambda}^{\mathrm{t}}\| \le \mu^{\mathrm{t}}\lambda^{n},\quad
|m^{\mathrm{tor}}| \le \mu^{\mathrm{tor}}\lambda^{n},\quad
\|\boldsymbol{m}^{\mathrm{rol}}\| \le \mu^{\mathrm{rol}}\lambda^{n}.
\label{eq:primal_cone}
\end{equation}
Here $\mu^*$s are  friction coefficients;   $\mu^{\mathrm{tor}}$ and $\mu^{\mathrm{roll}}$ have units of length (interpreted as effective contact-patch radius).

The corresponding dual cone constraints are defined in constraint-velocity space. Let 
$
(\boldsymbol{v}_{c},\boldsymbol{\omega}_{c}) {:=} \mathbf{J}\boldsymbol{v}
\label{eq:contact_velocities}
$
denote the \emph{relative}  linear and angular velocities between contact bodies, respectively. We decompose
$\boldsymbol{v}_{c} {:=}(v_{c}^{n},\boldsymbol{v}_{c}^{\mathrm{t}})$ into normal  velocity  $v_{c}^{n}\in \mathbb{R}$  and tangential slip velocity $\boldsymbol{v}_{c}^{\mathrm{t}} \in \mathbb{R}^{2}$. We also decompose   
$\boldsymbol{\omega}_{c} {:=}(\omega_{c}^{\mathrm{tor}},\boldsymbol{\omega}_{c}^{\mathrm{rol}})$ into   torsional angular velocity $\omega_{c}^{\mathrm{tor}}$ about the contact normal  and the rolling angular velocity $\boldsymbol{\omega}_{c}^{\mathrm{rol}} \in \mathbb{R}^{2}$  in the tangential plane.
Accordingly, define Jacobian blocks $\mathbf{J}_n,\mathbf{J}_{\mathrm{t}},\mathbf{J}_{\mathrm{tor}},\mathbf{J}_{\mathrm{rol}}$ such that
\begin{equation}
v_c^n=\mathbf{J}_n\boldsymbol{v},\,\,\,
\boldsymbol{v}_c^{\mathrm{t}}=\mathbf{J}_{\mathrm{t}}\boldsymbol{v},\,\,\,
\omega_c^{\mathrm{tor}}=\mathbf{J}_{\mathrm{tor}}\boldsymbol{v},\,\,\,
\boldsymbol{\omega}_c^{\mathrm{rol}}=\mathbf{J}_{\mathrm{rol}}\boldsymbol{v}.
\end{equation}
Then the dual-cone constraints corresponding to (\ref{eq:primal_cone}) are
\begin{equation}
\begin{aligned}
        \mu^{\mathrm{t}}\|\mathbf{J}_{\mathrm{t}}\boldsymbol{v}\|
&\le \mathbf{J}_n\boldsymbol{v}+{\phi}/{dt}\\
\mu^{\mathrm{tor}}|\mathbf{J}_{\mathrm{tor}}\boldsymbol{v}|
&\le \mathbf{J}_n\boldsymbol{v}+{\phi}/{dt}\\
\mu^{\mathrm{rol}}\|\mathbf{J}_{\mathrm{rol}}\boldsymbol{v}\|
&\le \mathbf{J}_n\boldsymbol{v}+{\phi}/{dt},
\end{aligned}
\label{eq:dual_cone}
\end{equation}
where $\phi$ is the signed gap distance between contact bodies. 
For notation simplicity, we  denote each dual cone constraint in (\ref{eq:dual_cone}) as $\mu^{\mathrm{s}}\|\mathbf{J}_{\mathrm{s}}\boldsymbol{v}\|
\le \mathbf{J}_n\boldsymbol{v}+{\phi}/{dt}$, with $\mathrm{s}\in\{\mathrm{t}, \mathrm{tor}, \mathrm{rol}\}$.

To facilitate computation, we approximate each quadratic dual-cone constraints  (\ref{eq:dual_cone}) with linear polyhedral constraints. For each $\mathrm{s}\in\{\mathrm{t}, \mathrm{tor}, \mathrm{rol}\}$, choose a symmetric set of $n_{\mathrm{s}}$ unit directions $\{\boldsymbol{d}^{(j)}_{\mathrm{s}}\}_{j=1}^{n_{\mathrm{s}}}$ spanning the corresponding subspace, and define the directional (row) Jacobian
$\mathbf{J}^{(j)}_{\mathrm{s}} := (\boldsymbol{d}^{(j)}_{\mathrm{s}})^\top \mathbf{J}_{\mathrm{s}}$.
This yields the linearized inequalities:
\begin{equation}
{\mathbf{\Tilde J}_{\mathrm{s}}}\boldsymbol{v}+ {\boldsymbol{\Tilde \phi}}/{dt}\geq0,
\quad  \mathrm{s}\in\{\mathrm{t}, \mathrm{tor}, \mathrm{rol}\}
\label{eq:dual_polyhedral}
\end{equation}
where 
\begin{equation}
{\mathbf{\Tilde J}_{\mathrm{s}}}:=\begin{bmatrix}
\mathbf{J}_{n}-\mu^{\mathrm{s}}\mathbf{J}^{(1)}_{\mathrm{s}}\\
\vdots\\
\mathbf{J}_{n}-\mu^{\mathrm{s}}\mathbf{J}^{(n_{\mathrm{s}})}_{\mathrm{s}}
\end{bmatrix},
\qquad
{\boldsymbol{\Tilde \phi}}:=\phi\,\mathbf{1}_{n_{\mathrm{s}}},
\label{eq:tilde_defs}
\end{equation}
and $\mathbf{J}^{(j)}_{\mathrm{s}}$ corresponds to the $j$-th polyhedral face, and $\ge 0$ is applied \emph{elementwise}.

\begin{figure}[h]
  \centering
  \includegraphics[width=0.90\linewidth]{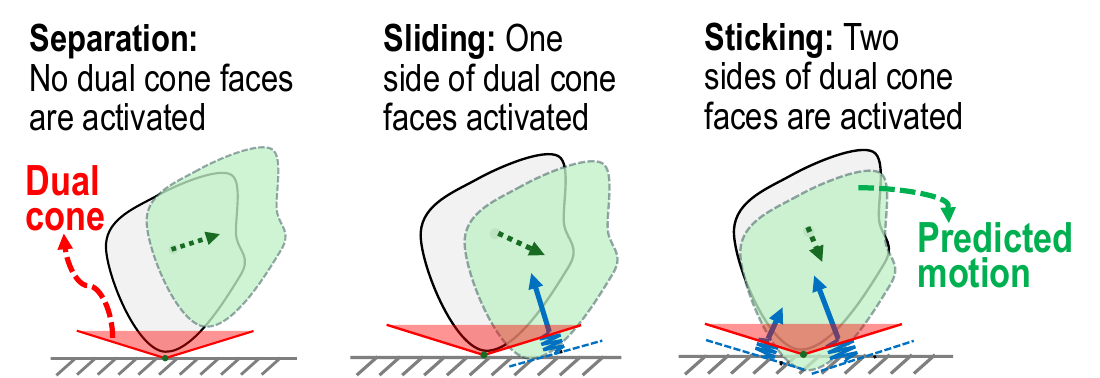}
  \caption{Different contact modes captured by ComFree-Sim.}
  \label{fig:contact_modes}
  \vspace{-10pt}
\end{figure}

The contact force/torque, based on  complementarity-free contact modeling, is analytically computed 
\begin{equation}
\label{eq:cf_wrench}
\boldsymbol{\lambda}_{\mathrm{s}}
{=}
\Big(-
\overbrace{K(\boldsymbol{q})
(\mathbf{\Tilde J}_{\mathrm{s}}\boldsymbol{v}_{\smooth}^+ dt
+\boldsymbol{\Tilde \phi})}^{\text{spring wrench}}
-
\overbrace{D(\boldsymbol{q})
(\mathbf{\Tilde J}_{\mathrm{s}}\boldsymbol{v}_{\smooth}^+)}^{\text{damping wrench}}
\Big)_+ .
\end{equation}
where $(x)_+ := \max(x,0)$ is applied element-wise. Here,
$\mathbf{\Tilde J}_{\mathrm{s}}\boldsymbol{v}_{\smooth}^+\,dt+\boldsymbol{\Tilde \phi}$
measures the violation (``penetration'') of the  dual-cone constraints under the  predicted smooth velocity $\boldsymbol{v}_{\smooth}^+$, and
$K(\boldsymbol{q})$ and $D(\boldsymbol{q})$ act as stiffness and damping gains in dual-cone space that map this violation to a stabilizing contact impulse.
With the above resolved contact forces/forces, the post contact velocity is written as 
\begin{equation}
\label{eq:vel_correction}
    \boldsymbol{v}^+=\boldsymbol{v}_{\smooth}^++\overbrace{
\boldsymbol{M}^{-1}\!\!\sum\nolimits_{\mathrm{s}\in\{\mathrm{t},\mathrm{tor},\mathrm{rol}\}}
\mathbf{\Tilde{J}}_{\mathrm{s}}^\trans\boldsymbol{\lambda}_{\mathrm{s}}dt
}^{\text{velocity correction}}
\end{equation}
Despite its simple form, this analytical update captures sticking, sliding, and separation via different activation patterns of $(\cdot)_+$ across polyhedral faces, as shown in Fig. \ref{fig:contact_modes}, while satisfying the primal Coulomb friction constraints  \eqref{eq:primal_cone} by construction; see  \cite{jin2024complementarity} for theoretical proof.

\vspace{-10pt}

\subsection{The Heuristics of Dual Cone Impedance}

In the above complementarity-free contact model, $K(\boldsymbol{q})$ and $D(\boldsymbol{q})$ are impedance parameters defined in the dual-cone space, and they play a central role in shaping the softness of resulting contact interaction. Following \cite{jin2024complementarity}, a  choice is to approximate the dual-cone impedance using a diagonal form,
\begin{equation}
K(\boldsymbol{q})\,dt + D(\boldsymbol{q}) \approx \frac{1}{dt}\Big(\mathbf{\Tilde J}\boldsymbol{M}^{-1}\mathbf{\Tilde J}^{\trans}\Big)^{-1}.
\end{equation}
However, evaluating the right-hand side exactly for every contact pair and time step can be computationally expensive. To reduce this overhead while preserving sufficient flexibility for practical simulation tuning, we introduce a simple heuristic parameterization of $K(\boldsymbol{q})$ and $D(\boldsymbol{q})$ that enables users to control the apparent ``stiffness'' of contact interaction with a small number of global parameters. In most use cases, this provides an effective ``one-size-fits-most'' setting. Moreover, ComFree-Sim also supports learning-based parameterization for $K(\boldsymbol{q})$ and $D(\boldsymbol{q})$ \cite{wang2026gswm}.
Specifically, we set
\begin{equation}\label{eq:k_d_setting}
K(\boldsymbol{q}) = k_{\text{user}}\,M(\phi)/dt,\quad
D(\boldsymbol{q}) = d_{\text{user}}\,M(\phi)/dt,
\end{equation}
where $k_{\text{user}}$ and $d_{\text{user}}$ are user-set global impedance parameters shared across all  contact pairs, and $M(\phi)$ is a constraint-space impedance that adapts to the signed distance between the current collision pair, i.e., bodies $i$ and $j$, 
Specifically,
\begin{equation}
M(\phi)=
\frac{r(|\phi|)}{1{-}r(|\phi|)}\cdot
\frac{{I}}{\tr(\mathbf{J}_i\boldsymbol{M}^{-1}\mathbf{J}_i^\trans){+}\tr(\mathbf{J}_j\boldsymbol{M}^{-1}\mathbf{J}_j^\trans)}.
\end{equation}
Here, ${I}$ is identity matrix, $r(|\phi|)\in(0,1)$ is a gap-dependent scaling factor, which is used in  MuJoCo backend solver \cite{todorov2012mujoco}:
\vspace{-15pt}
\begin{subequations}\label{eq:r_scaling}
\begin{align}
r(|\phi|) &= r_{\min}+(r_{\max}-r_{\min})\,\gamma(x), \quad x={|\phi|}/{w},\\
\gamma(x) &=
\begin{cases}
m\left(\dfrac{x}{m}\right)^{p}, & x<m,\\[6pt]
1-(1-m)\left(\dfrac{1-x}{1-m}\right)^{p}, & x\ge m,
\end{cases}
\end{align}
\end{subequations}
where the hyperparameters $(r_{\text{min}}, r_{\text{max}}, w, m, p)$ follow the identical  setting API as in MuJoCo. By default, they are set as $[0.9, 0.95, 0.001, 0.5, 2.0]$. This is to ensure ComFree-Sim share a MuJoCo-compatible interface.

\subsection{GPU Implementation for Contact Resolution}

A key advantage of ComFree-Sim is the  \emph{decoupling} nature of the  analytical contact resolution: it is independent across contact pairs and, under the polyhedral dual-cone approximation, separable across cone facets. This maps directly to GPU parallelism. Under Warp programming framework \cite{macklin2022warp}, ComFree-Sim launch kernels to (i) compute the smooth predicted velocity, (ii) solve dual-cone impulses in parallel over contacts and faces, (iii) accumulate generalized impulses, and (iv) apply  velocity correction, as shown in Algorithm~\ref{alg:comfree_gpu}.

\begin{algorithm}[h]
\caption{GPU-Parallelized ComFree-Sim Core}
\label{alg:comfree_gpu}
\footnotesize
\begin{algorithmic}
\REQUIRE State $(\boldsymbol{q},\boldsymbol{v})$, non-contact forces $(\boldsymbol{\tau},\boldsymbol{c})$, inertia $\boldsymbol{M}$, the detected contact set $\mathcal{C}$, where each contact $k\in\mathcal{C}$ contains signed gap $\phi_k$,  contact Jacobians  $\mathbf{J}_{n,k}$, $\{\mathbf{J}^{(j)}_{k,s}\}_{j=1}^{n_s}$;
friction coefficients $\{\mu^s_k\}_{s\in\{\mathrm{t},\mathrm{tor},\mathrm{rol}\}}$;
user-set global dual-cone  impedance $(k_\text{user},d_\text{user})$.
\ENSURE Post-contact velocity $\boldsymbol{v}^+$

\vspace{2pt}
\STATE \textbf{// Kernel I: Smooth prediction}
\STATE \hspace{2mm}\textbf{parallel over DoFs:} $\ \boldsymbol{v}^+_{\mathrm{smooth}} \leftarrow \boldsymbol{v} + \boldsymbol{M}^{-1}(\boldsymbol{\tau}-\boldsymbol{c})\,h$ \hfill (Eq.~(\ref{equ.smooth_vel}))

\vspace{2pt}
\STATE \textbf{// Kernel II: Per-contact, per-face dual-cone solve}
\STATE \hspace{2mm}\textbf{parallel over $k\in\mathcal{C}$, $s\in\{\mathrm{t},\mathrm{tor},\mathrm{rol}\}$,   $j=1..n_s$:}
\STATE \hspace{6mm} $\tilde{\mathbf{J}}_{k,s}^{(j)} \leftarrow \boldsymbol{J}^n_k - \mu_k^{s}\,\mathbf{J}^{(j)}_{k,s}$ \hfill (Eq.~(\ref{eq:tilde_defs}))
\STATE \hspace{6mm} $\tilde{\phi}_{k,s}^{(j)} \leftarrow \phi_k$
\STATE \hspace{6mm} $s_{k,s}^{(j)} \leftarrow \tilde{\mathbf{J}}_{k,s}^{(j)}\,\boldsymbol{v}^+_{\mathrm{smooth}}$ 
\STATE \hspace{6mm} $\boldsymbol{K}_{k,s}^{(j)}$, $\boldsymbol{D}_{k,s}^{(j)} \leftarrow $ $(k_\text{user},d_\text{user})$, $M(\phi_k)$ \hfill (Eq.~(\ref{eq:k_d_setting}))
\STATE \hspace{6mm} $\lambda_{k,s}^{(j)} \leftarrow \Big(\boldsymbol{K}_{k,s}^{(j)}\big(s_{k,s}^{(j)}dt + \tilde{\phi}_{k,s}^{(j)}\big) + \boldsymbol{D}_{k,s}^{(j)}\,s_{k,s}^{(j)}\Big)_{+}$ \hfill (Eq.~(\ref{eq:cf_wrench}))
\vspace{2pt}
\STATE \textbf{// Kernel III: Accumulate generalized impulse}
\STATE \hspace{2mm}\textbf{parallel over $(k,s,j)$:} $\ \boldsymbol{p} \mathrel{+}= \left(\tilde{\boldsymbol{J}}_{k,s}^{(j)}\right)^{\!\top}\lambda_{k,s}^{(j)}$
\vspace{2pt}
\STATE \textbf{// Kernel IV: Velocity correction}
\STATE \hspace{2mm}\textbf{parallel over DoFs:} $\ \boldsymbol{v}^+ \leftarrow \boldsymbol{v}^+_{\mathrm{smooth}} + \boldsymbol{M}^{-1}\boldsymbol{p}\,dt$ \hfill (Eq.~(\ref{eq:vel_correction}))
\RETURN $\boldsymbol{v}^+$
\end{algorithmic}
\end{algorithm}

\section{GPU Simulation Benchmarking}
To  evaluate ComFree-Sim, we benchmark simulation fidelity, stability, and efficiency against MJWarp~\cite{mujoco_warp}, a state-of-the-art GPU simulator. Specifically, we measure penetration depth, torsional/rolling friction behaviors, stability, runtime scaling with contact count, and parallel throughput. To isolate  contact-resolution backend, both simulators use the same collision detection and time integration ~\cite{mujoco_warp}. All tests run on an AMD 32-core CPU with an NVIDIA RTX 4090 GPU.

\subsection{Penetration Depth}
\begin{figure}[!htbp]
    \centering
    \begin{subfigure}{0.35\linewidth}
        \includegraphics[width=\linewidth]{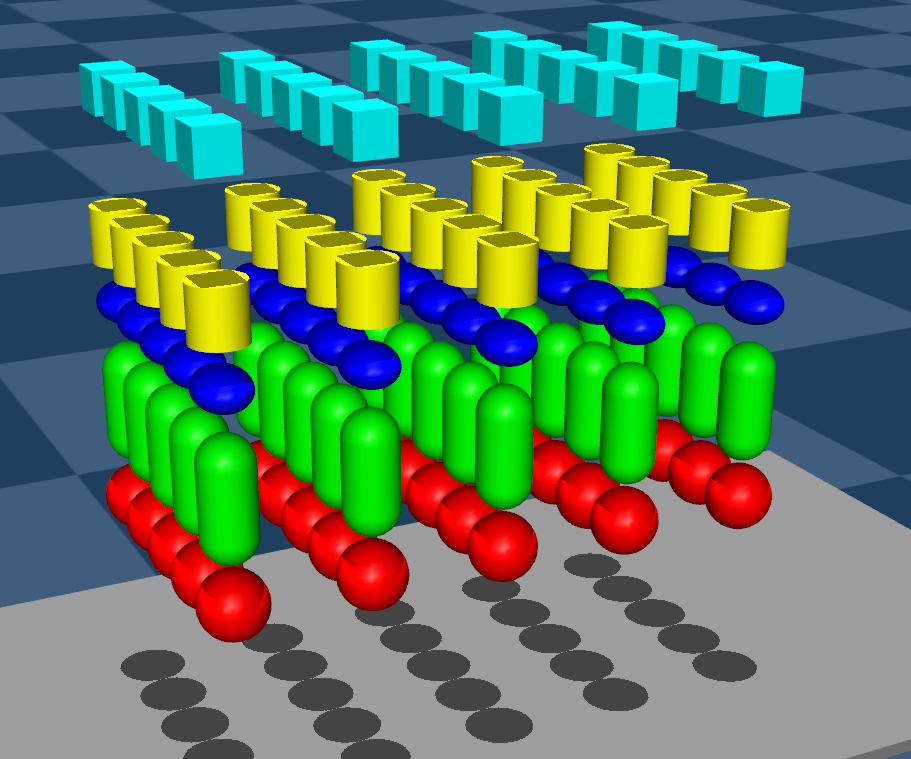}
        \caption{Test Env.}
    \label{fig:penetration_env}
    \end{subfigure}
    \hfill
    \begin{subfigure}{0.6\linewidth}
        \centering
        \scriptsize
        \begin{tabular}{lll}
        \toprule
        Engine & $k_\text{user},d_\text{user}$ & Depth (mm) \\
        \midrule
        Mujoco-Warp & N/A & 1.7 ± 4.9 \\
        ComFree-Sim & 0.1, 0.001 & 3.9 ± 6.9 \\
        ComFree-Sim & 0.1, 0.005 & 3.8 ± 5.7 \\
        ComFree-Sim & 0.3, 0.001 & 1.6 ± 3.3 \\
        ComFree-Sim & 0.3, 0.005 & 1.4 ± 2.5 \\
        ComFree-Sim & 0.5, 0.001 & 1.0 ± 2.1 \\
        ComFree-Sim & 0.5, 0.005 & 0.9 ± 1.5 \\
        \bottomrule
        \end{tabular}
        \caption{Penetration depths.}
    \label{tbl:depths}
    \end{subfigure}
\caption{Penetration-depth benchmark in a collision-rich drop test: (a) stacked arrays of  primitives released to free-fall onto a flat box; (b) mean$\pm$std penetration depth over all detected contacts. In all tests,  simulation timestep $dt=0.002$.}
    \vspace{-10pt}
\end{figure}

Penetration depth indicates rigid-body fidelity: larger interpenetration implies a stronger violation of rigidity. We compare MJWarp and ComFree-Sim in a collision-rich drop test where five $5\times5$ arrays of convex primitives (cubes, cylinders, ellipsoids, capsules, and spheres; average size $\approx$5\,cm) are stacked (Fig.~\ref{fig:penetration_env}) and dropped onto a flat box, inducing dense contacts. At each step, we record the penetration depth of all detected contact pairs; after 1000 steps, we report the mean and standard deviation for MJWarp and multiple ComFree-Sim settings (Fig.~\ref{tbl:depths}). Overall, penetration increases with $k_{\text{user}}$ and decreases with $d_{\text{user}}$; with appropriate tuning, ComFree-Sim achieves comparable or lower penetration than MJWarp.

\vspace{-5pt}
\subsection{Torsional and Rolling Friction Modelling}

\begin{figure}[!htbp]
    \centering
    \begin{subfigure}[b]{0.25\linewidth}
        \centering
        \includegraphics[width=\linewidth]{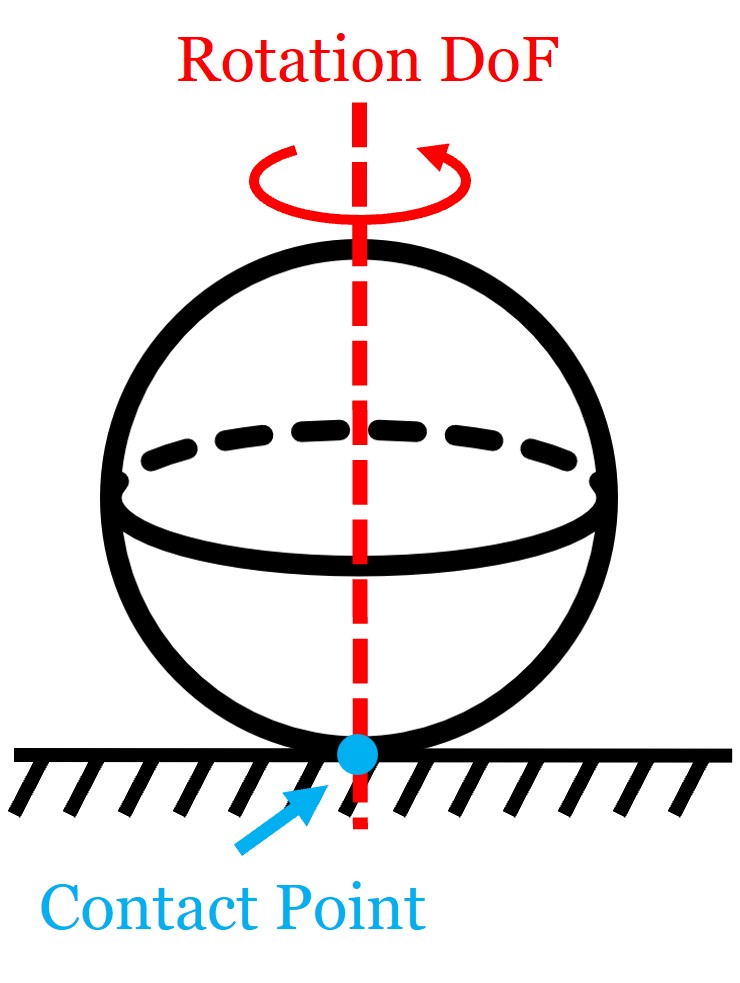}
        \vspace{-20pt}
        \caption{Torsional}
    \label{fig:torsion_env}
    \end{subfigure}
    \hfill
    \begin{subfigure}[b]{0.3\linewidth}
        \centering
        \includegraphics[width=\linewidth]{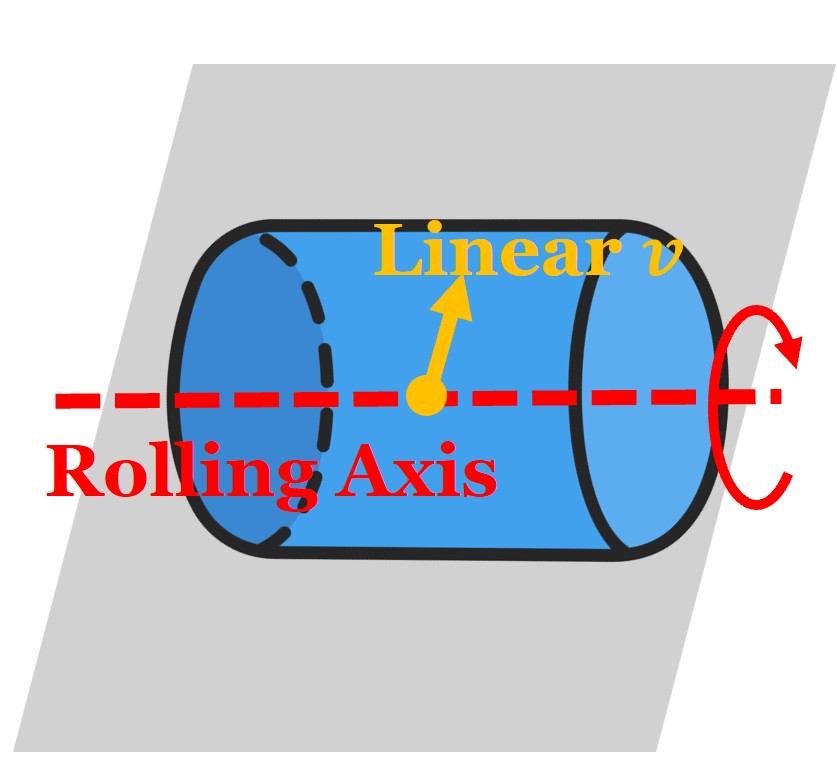}
        \caption{Rolling}
        \label{fig:rolling_env}
    \end{subfigure}
    \hfill
    \begin{subfigure}[b]{0.3\linewidth}
        \centering
        \includegraphics[width=\linewidth]{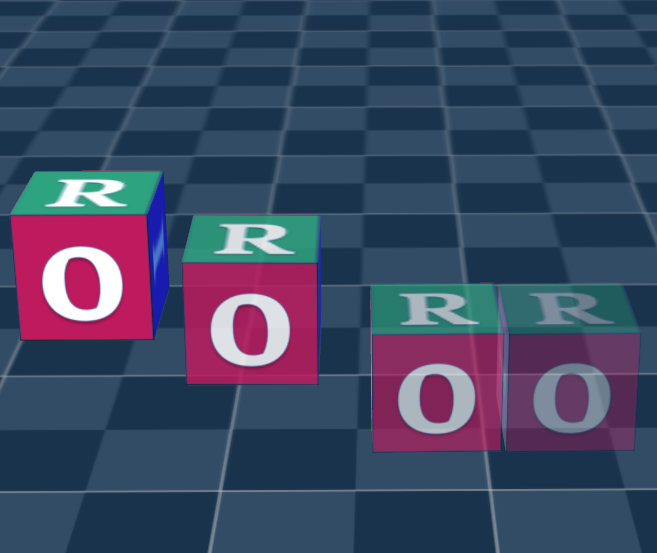}
        \caption{Stability}
        \label{fig:stability_env}
    \end{subfigure}
\caption{Isolated benchmark environments.}   
\label{fig:torsion}
\vspace{-10pt}
\end{figure}

\begin{figure}[!htbp]
    \centering
    \begin{subfigure}{0.48\linewidth}
        \centering
        \includegraphics[width = \linewidth]{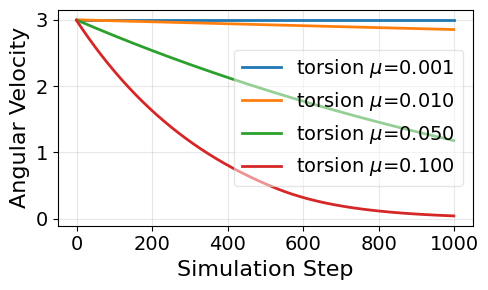}
        \caption{Torsional: angular velocity.}
    \label{fig:torsion_plot}
    \end{subfigure}
    \hfill
    \begin{subfigure}{0.48\linewidth}
        \centering
        \includegraphics[width = \linewidth]{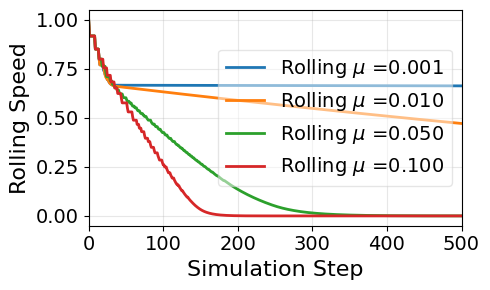}
        \caption{Rolling: linear velocity.}
        \label{fig:rolling_plot}
    \end{subfigure}
\caption{Results of torsional and rolling friction test.}   
\label{fig:friction_plots}
\vspace{-10pt}
\end{figure}

We evaluate torsional and rolling friction with two controlled contact tests that isolate each effect. For torsional friction, a sphere in contact with a plane is constrained to rotate only about  $z$-axis (Fig.~\ref{fig:torsion_env}), and we measure angular-velocity decay under varying torsional friction coefficients (Fig.~\ref{fig:torsion_plot}). For rolling friction, a cylinder rolls freely on a plane (Fig.~\ref{fig:rolling_env}), and we report the decay of its COM velocity under varying rolling friction coefficients (Fig.~\ref{fig:rolling_plot}). The monotone decay trends and their response to the corresponding coefficients confirm  ComFree-Sim captures torsional/rolling dissipation with  consistent contact dynamics.

\subsection{Simulation Stability}

\begin{figure}[!htbp]
    \centering
    \includegraphics[width=0.99\linewidth]{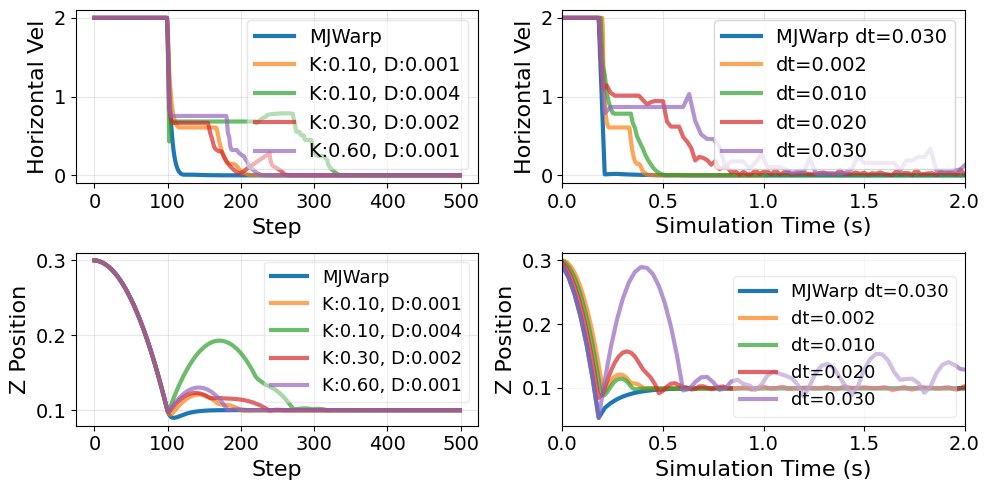}
\caption{Numerical stability test: horizontal speed and vertical position $z$ of free-fall-and-sliding cube over time. Left: ComFree-Sim with varying   $(k_{\text{user}},d_{\text{user}})$  at $d t{=}0.002$s. Right: ComFree-Sim under varying timestep $dt$.}
    \label{fig:stability_plot}
    \vspace{-15pt}
\end{figure}

To assess numerical stability, we run a free-fall-and-sliding test where a cube is dropped onto a flat plane with initial linear velocity $(2.0,0,0)$ and angular velocity $(0.1,0.1,0.1)$ under gravity and friction (Fig.~\ref{fig:stability_env}). Since friction should dissipate  kinetic energy and drive the cube to rest, we track the horizontal speed and $z$ (vertical) position over time (Fig.~\ref{fig:stability_plot}). We sweep ComFree impedance settings ($k_{\text{user}}, d_{\text{user}}$) and time steps $dt$. Across a wide range of parameters, ComFree-Sim exhibits consistent, monotone horizontal-velocity decay without spurious drift or growth, indicating low sensitivity to moderate impedance variations. ComFree-Sim remains stable  at a moderately large time step (e.g., $dt{=}0.02$\,s), but typically benefits from smaller $dt$ than MJWarp; unless otherwise noted, we use $dt{=}0.002$s in all benchmarks.

\subsection{Runtime Performance w.r.t. Contact Numbers}
\begin{figure}[!htbp]
\centering
\includegraphics[width=0.6\columnwidth]{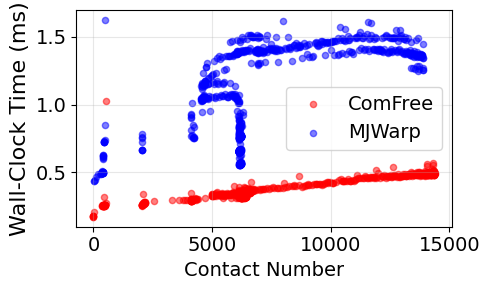}
\caption{Runtime scaling benchmark. Y:  wall-clock step time.}
\label{fig:tp}
\vspace{-15pt}
\end{figure}

We evaluate runtime scaling by measuring the \emph{wall-clock time per full simulation step} as a function of the number of contacts in the scene. This metric directly reflects computational efficiency in contact-rich simulation.
We instantiate 512 parallel environments. In each environment, three $2\times2$ arrays of convex primitives (cubes, capsules, and spheres) are stacked top-down (similar to Fig.~\ref{fig:penetration_env}). To introduce stochasticity, we perturb all bodies with small initial linear velocities sampled from $\mathcal{N}(0,10^{-3})$. We then run 750 simulation steps for MJWarp and ComFree-Sim ($k_{\text{user}}=0.1$, $d_{\text{user}}=0.001$, and $dt=0.002$). At each step, we record the total contacts (aggregated over all environments) and the full-step wall-clock time (excluding rendering), including collision detection, contact resolution, and integration; Fig.~\ref{fig:tp} plots wall-clock time  step time versus contact count.

ComFree-Sim achieves around $3\times$ faster simulation speed and exhibits an approximately \emph{linear} step-time scaling with contact count, consistent with its analytical, per-contact decoupled update and GPU parallelism. MJWarp shows \emph{nonlinear} (often superlinear) step-time growth and higher variance at comparable contact counts. Overall, ComFree-Sim provides higher throughput and more predictable scaling in dense contact scenes.


\subsection{Throughput for Parallel Simulation}

\begin{figure}[!htbp]
    \centering
    \begin{subfigure}[b]{0.3\linewidth}
        \centering
        \includegraphics[width=\linewidth]{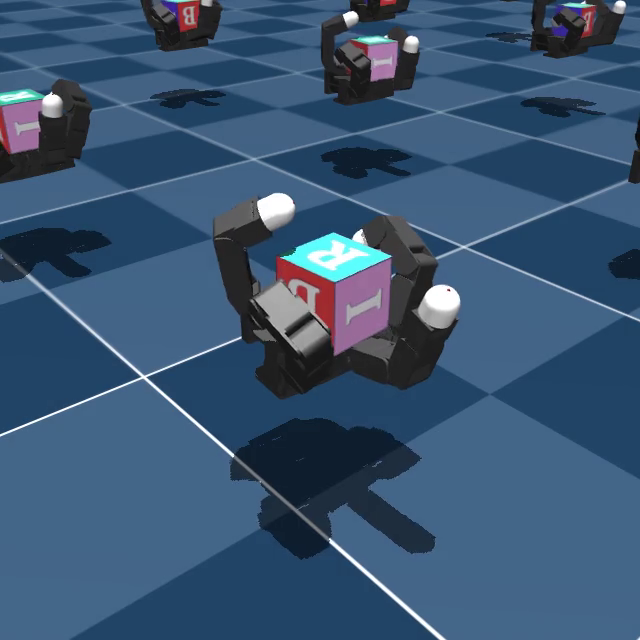}
        \caption{Parallel Env.}
        \label{fig:allegro-parallel}
    \end{subfigure}
    \hfill
    \begin{subfigure}[b]{0.65\linewidth}
        \centering
        \includegraphics[width=\linewidth]{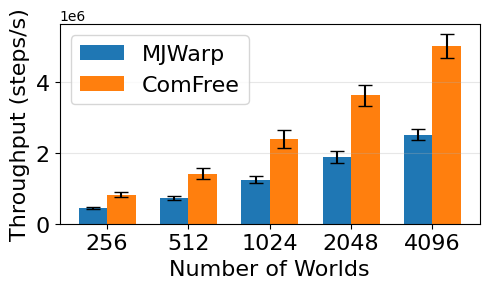}
        \vspace{-20pt}
        \caption{Throughput of parallel simulation.}
        \label{fig:parallel}
    \end{subfigure}
\caption{Allegro hand throughput test.}   
\label{fig:allegro}
\vspace{-15pt}
\end{figure}

We benchmark simulation throughput versus the number of parallel environments on an Allegro in-hand cube grasping scene (Fig.~\ref{fig:allegro-parallel}). Starting from a stable grasp, we run 1500 simulation steps with 256/512/1024/2048/4096 environments and report the mean$\pm$std throughput, defined as \textbf{single-environment} steps per second. To sustain rich robot--object contact, every 50 steps we apply a random joint-position command within $(-0.5,0.5)$ around the initial configuration while preserving the grasp. As shown in Fig.~\ref{fig:parallel}, ComFree-Sim achieves nearly $2\times$ the throughput of MJWarp, highlighting its advantage for large-scale sampling-based control and robot learning.

\section{ComFree-Sim for Real-Time MPC for Real-World Dexterous Manipulation}

\begin{table*}[t]
\renewcommand{\arraystretch}{1.3} 
\centering
\caption{Control and runtime performance for real-world in-hand dexterous manipulation with LEAP Hand}
\label{table_combined_results}
\footnotesize
\setlength{\tabcolsep}{5.75pt} 
\begin{tabular}{lcccccccccccc}
\toprule
& \multicolumn{2}{c}{Pos Error [m]} & \multicolumn{2}{c}{Quat Error} & \multicolumn{2}{c}{MPPI Time [ms]} & \multicolumn{2}{c}{Task completion time [s]} & \multicolumn{2}{c}{Success Rate (\%)} \\
\cmidrule(lr){2-3} \cmidrule(lr){4-5} \cmidrule(lr){6-7} \cmidrule(lr){8-9} \cmidrule(lr){10-11} \cmidrule(lr){12-13}
\makecell{Object} & ComFree & MJWarp & ComFree & MJWarp & ComFree & MJWarp & ComFree & MJWarp & ComFree & MJWarp \\

\midrule

\makecell{Cube\\(Yaw)} & \makecell{\textbf{0.0106} \\ \textbf{$\pm$0.0028}} & \makecell{0.0119 \\ $\pm$0.0041} & \makecell{\textbf{0.0231} \\ \textbf{$\pm$0.0081}} & \makecell{0.0236 \\ $\pm$0.0103} & \makecell{\textbf{28.2308} \\ \textbf{$\pm$2.0053}} & \makecell{54.5961 \\ $\pm$5.5158} & \makecell{\textbf{42.4261} \\ \textbf{$\pm$21.3724}} & \makecell{60.6831 \\ $\pm$40.9661} & \textbf{80.00} & 65.00 \\

\midrule

\makecell{Cube\\(Roll/Pitch)} & \makecell{\textbf{0.0087} \\ \textbf{$\pm$0.0036}} & \makecell{0.0114 \\ $\pm$0.0032} & \makecell{\textbf{0.0250} \\ \textbf{$\pm$0.0084}} & \makecell{0.0291 \\ $\pm$0.0054} & \makecell{\textbf{27.7969} \\ \textbf{$\pm$1.0890}} & \makecell{56.4415 \\ $\pm$5.5757} & \makecell{\textbf{20.1756} \\ \textbf{$\pm$12.0413}} & \makecell{74.3734 \\ $\pm$44.2553} & \textbf{55.00} & 30.00 \\

\midrule

\makecell{Duck\\(Yaw)} & \makecell{0.0150 \\ $\pm$0.0020} & \makecell{\textbf{0.0116} \\ \textbf{$\pm$0.0028}} & \makecell{0.0255 \\ $\pm$0.0066} & \makecell{\textbf{0.0186} \\ \textbf{$\pm$0.0104}} & \makecell{\textbf{18.2271} \\ \textbf{$\pm$0.7618}} & \makecell{41.8674 \\ $\pm$9.3805} & \makecell{\textbf{28.7974} \\ \textbf{$\pm$18.4190}} & \makecell{49.4049 \\ $\pm$29.3045} & \textbf{50.00} & 25.00 \\

\midrule

\makecell{Duck\\(Roll/Pitch)} & \makecell{0.0183 \\ $\pm$0.0012} & \makecell{\textbf{0.0180} \\ \textbf{$\pm$0.0000}} & \makecell{\textbf{0.0191} \\ \textbf{$\pm$0.0036}} & \makecell{0.0195 \\ $\pm$0.0000} & \makecell{\textbf{13.9000} \\ \textbf{$\pm$5.8266}} & \makecell{49.2301 \\ $\pm$0.0000} & \makecell{24.8061 \\ $\pm$15.4174} & \makecell{\textbf{7.6724} \\ \textbf{$\pm$0.0000}} & \textbf{20.00} & 10.00 \\

\midrule

\makecell{SpamTin\\(Yaw)} & \makecell{\textbf{0.0161} \\ \textbf{$\pm$0.0013}} & \makecell{0.0164 \\ $\pm$0.0011} & \makecell{0.0194 \\ $\pm$0.0085} & \makecell{\textbf{0.0172} \\ \textbf{$\pm$0.0036}} & \makecell{\textbf{20.0710} \\ \textbf{$\pm$0.8337}} & \makecell{52.4135 \\ $\pm$2.4810} & \makecell{\textbf{29.3871} \\ \textbf{$\pm$18.3080}} & \makecell{53.0317 \\ $\pm$21.1797} & \textbf{65.00} & 15.00 \\

\midrule

\makecell{SpamTin\\(Roll/Pitch)} & \makecell{0.0178 \\ $\pm$0.0009} & \makecell{\textbf{0.0174} \\ \textbf{$\pm$0.0005}} & \makecell{\textbf{0.0198} \\ \textbf{$\pm$0.0050}} & \makecell{0.0272 \\ $\pm$0.0029} & \makecell{\textbf{19.2457} \\ \textbf{$\pm$0.2715}} & \makecell{68.7749 \\ $\pm$5.8720} & \makecell{36.0421 \\ $\pm$30.5504} & \makecell{\textbf{21.7259} \\ \textbf{$\pm$6.0693}} & \textbf{40.00} & 30.00 \\

\midrule

\makecell{Cylinder\\(Yaw)} & \makecell{0.0147 \\ $\pm$0.0037} & \makecell{\textbf{0.0103} \\ \textbf{$\pm$0.0005}} & \makecell{\textbf{0.0221} \\ \textbf{$\pm$0.0094}} & \makecell{0.0241 \\ $\pm$0.0042} & \makecell{\textbf{20.7444} \\ \textbf{$\pm$0.9223}} & \makecell{39.2304 \\ $\pm$2.3594} & \makecell{34.1209 \\ $\pm$25.4092} & \makecell{\textbf{27.7529} \\ \textbf{$\pm$18.7242}} & \textbf{50.00} & 10.00 \\

\midrule

\makecell{Cylinder\\(Roll/Pitch)} & \makecell{\textbf{0.0145} \\ \textbf{$\pm$0.0024}} & \makecell{0.0162 \\ $\pm$0.0000} & \makecell{0.0183 \\ $\pm$0.0068} & \makecell{\textbf{0.0040} \\ \textbf{$\pm$0.0000}} & \makecell{\textbf{21.1287} \\ \textbf{$\pm$1.2188}} & \makecell{34.0165 \\ $\pm$0.0000} & \makecell{\textbf{27.1724} \\ \textbf{$\pm$19.8587}} & \makecell{62.2975 \\ $\pm$0.0000} & \textbf{50.00} & 10.00 \\
\midrule
\end{tabular}
\vspace{-10pt}
\end{table*}


We deploy ComFree-Sim as the predictive model in a real-time MPC loop for multi-fingered in-hand object reorientation on a physical LEAP Hand \cite{shaw2023leaphand}. The goal is to reorient objects to  target poses, evaluating ComFree-Sim’s closed-loop real-time performance and model gap  under real-world contact-rich control. The MPC is formulated as 
\begin{equation}
\label{eq:mpc}
\begin{aligned}
\min_{\mathbf{U}:=(\boldsymbol{u}_{0:H-1})}\quad
& J(\mathbf{U})=\sum\nolimits_{t=0}^{H-1} c(\boldsymbol{u}_{t},\boldsymbol{x}_{t}) + V(\boldsymbol{x}_H)\\
\text{s.t.}\quad
& \boldsymbol{x}_{t+1}=\textrm{ComFree-Sim}(\boldsymbol{x}_{t}, \boldsymbol{u}_{t}), \\
& \boldsymbol{x}_0=\boldsymbol{x}^{\text{env}} \quad t=0,1,\ldots,H-1.
\end{aligned}
\end{equation}
Here, $\boldsymbol{x}_t$ denotes the system state and $\boldsymbol{u}_t$ the control input at time step $t$. We set the running and terminal costs as
\begin{equation}
    \begin{aligned}
        c := & \, \omega_{1} c_{\text{quat}}(\boldsymbol{x}) + \omega_{2} p_{x}(\boldsymbol{x}) + \omega_{3} p_{y}(\boldsymbol{x}) + \omega_{4} p_{z}(\boldsymbol{x}) \\
        & + \omega_{5} c_{\text{contact}}(\boldsymbol{x}) + \omega_{6} c_{\text{joint}}(\boldsymbol{x}) + \Omega \mathbb{I}_{\text{fallen}}(\boldsymbol{x}) \\
        V &:= \phi_{1} \|\mathbf{p}^{\text{obj}} - \mathbf{p}_{\text{target}}\|^2 + \phi_{2} (1-(\mathbf{q}_{\text{target}}^{\top} \mathbf{q}^{\text{obj}})^2)
    \end{aligned}
\end{equation}
where $c_{\text{quat}} {:=} 1{-}(\mathbf{q}_{\text{target}}^{\top}\mathbf{q}^{\text{obj}})^2$ is  orientation error; $(p_x,p_y,p_z)$ are the absolute position errors; $c_{\text{contact}} {:=} \sum_{i=1}^{4}\|\mathbf{p}^{\text{obj}}-\mathbf{p}^{\text{f/t}_i}\|^2$ penalizes fingertip-to-object-center distance; $c_{\text{joint}} {:=} \|\boldsymbol{q}_{\text{robot}}-\boldsymbol{q}_{\text{ref}}\|^2$ penalizes deviation from a home pose; and $\mathbb{I}_{\text{fallen}}=1$ if $p_z<0.05$. We tune $\omega_k$ and $(\phi_1,\phi_2)$ based on the object mass and geometry.

Throughout, we set the ComFree-Sim  $dt{=}0.004$\,s and horizon $H{=}48$. We solve \eqref{eq:mpc} using MPPI \cite{williams2017model} with $N=256$ samples, temperature $\lambda=2\times10^{-3}$, and sampling standard deviation $0.02$. We use incremental position control and clip actions to $[-0.1,0.1]$. For a controlled comparison, we replace ComFree-Sim with MJWarp while keeping all other settings identical.

\subsection{Hardware Setup}
\vspace{-10pt}
\begin{figure}[!htbp]
\centering
\includegraphics[width=1.0\columnwidth]{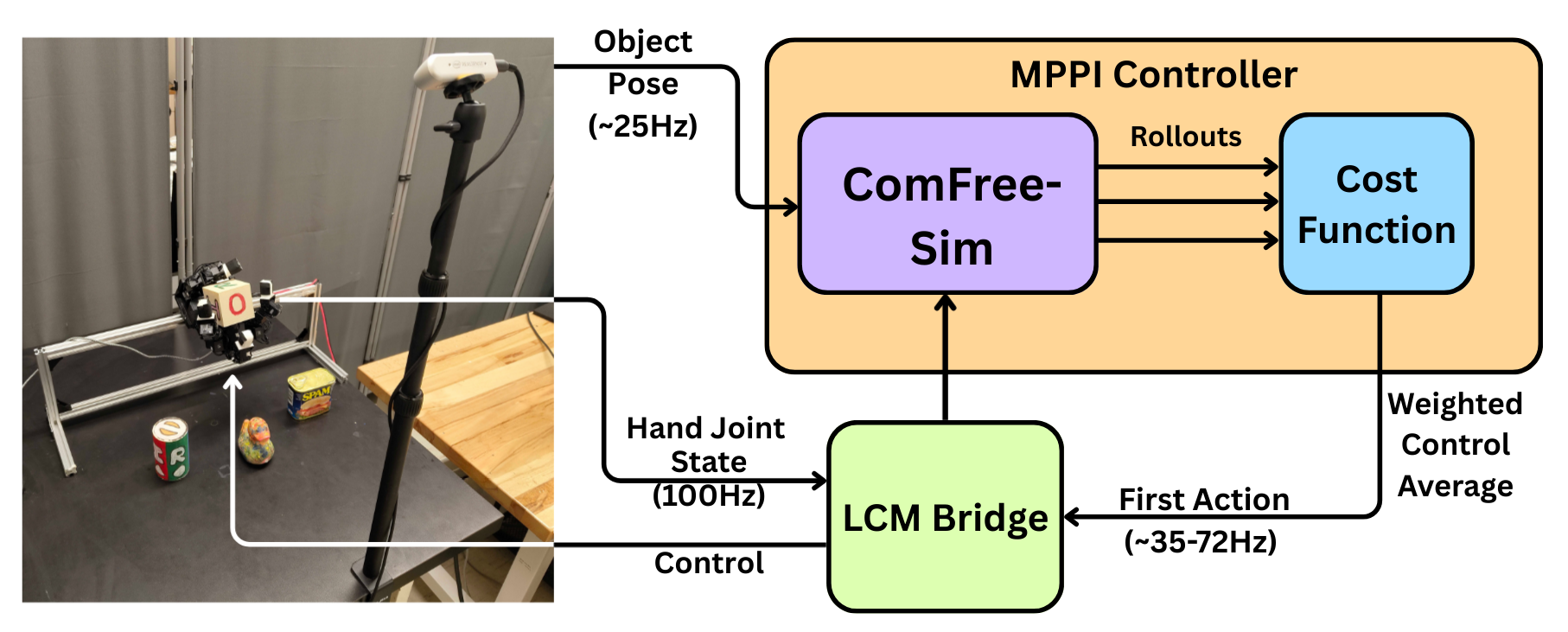}
\caption{Hardware setup and system diagram.}
\label{fig:setup}
\vspace{-5pt}
\end{figure}
The real-world LEAP Hand platform for dexterous manipulation is shown in Fig. \ref{fig:setup}. The experiment covers objects with diverse (convex or nonconvex) shapes, including Cube, Duck, SpamTin, and Cylinder. Object   pose is estimated using an Intel RealSense D435i RGB-D camera mounted above the workspace. We use  FoundationPose \cite{wen2024foundationposeunified6dpose} to perform 6D object pose tracking. The estimated object pose is streamed to the controller at $\sim25$ Hz via LCM\cite{5649358}. Robot joint states are read at 100 Hz via the LEAP Hand API and sent to the controller through LCM. ComFree-MPPI Controller receives robot joint states and object pose. We set 
$k_{\text{user}}=0.1$ and $d_{\text{user}}=0.002$ 
for all objects. At each control step, MPPI samples $N$ control sequences and rolls them out over a finite horizon using ComFree simulator. A weighted average control sequence is computed and first control input is applied in a receding-horizon manner as joint position commands. The closed-loop control frequency ranges between 35–72 Hz, depending on the number of contacts. The hand is mounted on a rigid steel rail at $\sim28^{\circ}$ inclination to provide a stable base for in-hand manipulation experiments. All algorithms run on an AMD 32-core CPU with  NVIDIA  4090 GPUs.


\begin{table*}[h]
\renewcommand{\arraystretch}{1.3}
\centering
\caption{Comparison of Motion Retargeting Performance Across Tasks}
\label{tab:retargeting}
\scriptsize
\setlength{\tabcolsep}{4pt}

\begin{tabular}{lcccccccccccc}
\toprule
& \multicolumn{2}{c}{Pos. Error} 
& \multicolumn{2}{c}{Quat. Error} 
& \multicolumn{2}{c}{Joint Error} 
& \multicolumn{2}{c}{Obj. Error} 
& \multicolumn{2}{c}{Obj. Quat. Error} 
& \multicolumn{2}{c}{MPPI. Time (s)} \\

\cmidrule(lr){2-3} 
\cmidrule(lr){4-5} 
\cmidrule(lr){6-7}
\cmidrule(lr){8-9}
\cmidrule(lr){10-11}
\cmidrule(lr){12-13}

Task 
& ComFree & MJWarp 
& ComFree & MJWarp 
& ComFree & MJWarp 
& ComFree & MJWarp 
& ComFree & MJWarp 
& ComFree & MJWarp \\

\midrule

Cartwheels
& \makecell{0.037 \\ $\pm$0.002}
& \makecell{\textbf{0.033} \\ \textbf{$\pm$0.001}}
& \makecell{0.100 \\ $\pm$0.004}
& \makecell{\textbf{0.074} \\ \textbf{$\pm$0.001}}
& \makecell{0.692 \\ $\pm$0.008}
& \makecell{\textbf{0.682} \\ \textbf{$\pm$0.007}}
& -- & --
& -- & --
& \makecell{\textbf{0.529} \\ \textbf{$\pm$0.422}}
& \makecell{0.705 \\ $\pm$0.490} \\

\midrule

Martial Arts
& \makecell{0.047 \\ $\pm$0.003}
& \makecell{\textbf{0.028} \\ \textbf{$\pm$0.001}}
& \makecell{0.118 \\ $\pm$0.005}
& \makecell{\textbf{0.076} \\ \textbf{$\pm$0.002}}
& \makecell{0.830 \\ $\pm$0.007}
& \makecell{\textbf{0.794} \\ \textbf{$\pm$0.007}}
& -- & --
& -- & --
& \makecell{\textbf{0.589} \\ \textbf{$\pm$0.336}}
& \makecell{0.774 \\ $\pm$0.492} \\

\midrule

Dance
& \makecell{0.037 \\ $\pm$0.001}
& \makecell{\textbf{0.033} \\ \textbf{$\pm$0.000}}
& \makecell{0.094 \\ $\pm$0.003}
& \makecell{\textbf{0.069} \\ \textbf{$\pm$0.000}}
& \makecell{0.881 \\ $\pm$0.003}
& \makecell{\textbf{0.871} \\ \textbf{$\pm$0.001}}
& -- & --
& -- & --
& \makecell{\textbf{0.481} \\ \textbf{$\pm$0.367}}
& \makecell{0.646 \\ $\pm$0.428} \\

\midrule

Getup
& \makecell{0.036 \\ $\pm$0.001}
& \makecell{\textbf{0.032} \\ \textbf{$\pm$0.000}}
& \makecell{0.088 \\ $\pm$0.003}
& \makecell{\textbf{0.068} \\ \textbf{$\pm$0.001}}
& \makecell{0.705 \\ $\pm$0.003}
& \makecell{\textbf{0.682} \\ \textbf{$\pm$0.003}}
& -- & --
& -- & --
& \makecell{\textbf{0.473} \\ \textbf{$\pm$0.388}}
& \makecell{0.572 \\ $\pm$0.342} \\

\midrule

PushBox
& \makecell{0.103 \\ $\pm$0.017}
& \makecell{\textbf{0.080} \\ \textbf{$\pm$0.005}}
& \makecell{0.079 \\ $\pm$0.010}
& \makecell{\textbf{0.057} \\ \textbf{$\pm$0.002}}
& \makecell{0.246 \\ $\pm$0.022}
& \makecell{\textbf{0.196} \\ \textbf{$\pm$0.004}}
& \makecell{0.169 \\ $\pm$0.082}
& \makecell{\textbf{0.098} \\ \textbf{$\pm$0.010}}
& \makecell{0.260 \\ $\pm$0.146}
& \makecell{\textbf{0.161} \\ \textbf{$\pm$0.018}}
& \makecell{1.171 \\ $\pm$1.189}
& \makecell{\textbf{0.719} \\ \textbf{$\pm$0.708}} \\

\bottomrule
\end{tabular}
\end{table*}

\subsection{Results and Analysis}

\subsubsection{Metrics and Results}
We evaluate five metrics: \textbf{Success rate}, a trial is successful if position error $<0.02$\,m and quaternion error $<0.04$;
\textbf{MPPI time}, wall-clock compute time per MPC control step;
\textbf{Position error}, $\|\mathbf{p}^{\text{obj}}-\mathbf{p}_{\text{target}}\|$;
\textbf{Quaternion error}, $1{-}(\mathbf{q}_{\text{target}}^{\top}\mathbf{q}^{\text{obj}})^2$;
and \textbf{Task completion time}, elapsed real time to reach the success condition.
For each object/target, we run 20 independent trials; success rate and MPPI time are reported over all trials, while position and quaternion errors are computed over successful trials only.



\subsubsection{Analysis}
Overall, ComFree-Sim achieves manipulation accuracy comparable to MJWarp: MJWarp yields lower raw error on some tasks (e.g., Duck-Yaw, Cylinder-Yaw), while ComFree-Sim matches or improves others (e.g., Cube-Roll/Pitch, Cylinder-Roll/Pitch). Crucially, ComFree-Sim reduces per-step MPPI compute time for \emph{all} object--task pairs, with $\sim$2.4$\times$ average speedup (median $\sim$2.2$\times$), which consistently boosts closed-loop success rates (+27 percentage points on average across eight benchmarks). With sub-30\,ms MPPI step times (13.9--28.2\,ms in Table~\ref{table_combined_results}), ComFree-Sim enables $\sim$35--72\,Hz control on hardware, making it a strong backend for high-frequency, contact-rich MPC.

\section{ComFree-Sim for Dynamic-Aware Retargeting for Agile Locomotion}

We evaluate ComFree-Sim on model-based, dynamics-aware motion retargeting using SPIDER~\cite{pan2025spider}, which optimizes the control sequence $\mathbf{U}$ in~\eqref{eq:mpc} via MPPI with an annealed noise schedule~\cite{xue2025full}. The optimal control sequence $\mathbf{U}^*$ is applied to a  simulator in a receding-horizon manner to generate the full trajectory, ensuring physical plausibility. The original SPIDER method ~\cite{pan2025spider} performs optimization  in MJWarp and generates  full trajectories in MuJoCo-CPU~\cite{mujoco_warp}.

\begin{figure}[!htbp]
    \centering
    \begin{subfigure}[c]{0.32\linewidth}
        \centering
        \includegraphics[width=\linewidth]{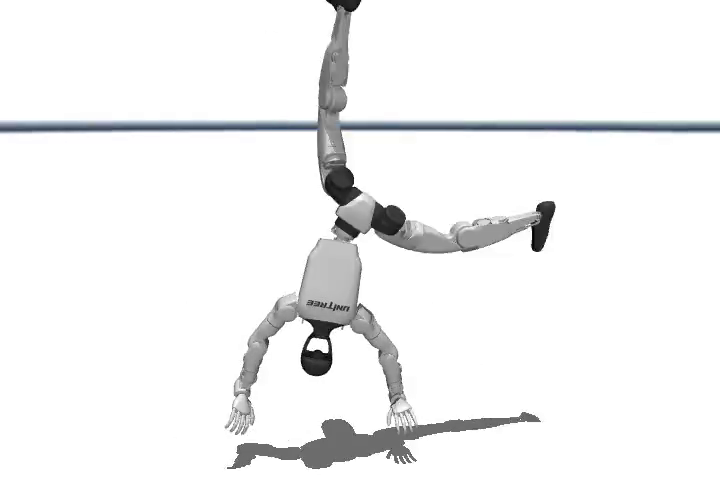}
        \caption{Cartwheels}
    \end{subfigure}
    \hfill
    \begin{subfigure}[c]{0.32\linewidth}
        \centering
        \includegraphics[width=\linewidth]{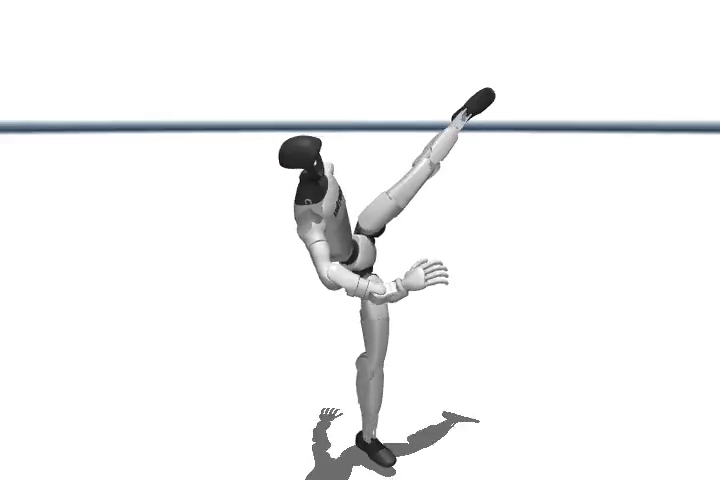}
        \caption{Martial Arts}
    \end{subfigure}
    \hfill
    \begin{subfigure}[c]{0.32\linewidth}
        \centering
        \includegraphics[width=\linewidth]{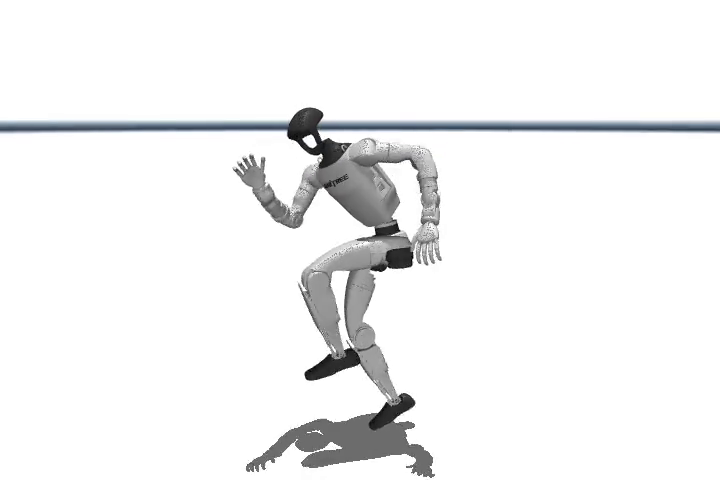}
        \caption{Dance}
    \end{subfigure}
    \\
    \begin{subfigure}[c]{0.32\linewidth}
        \centering
        \includegraphics[width=\linewidth]{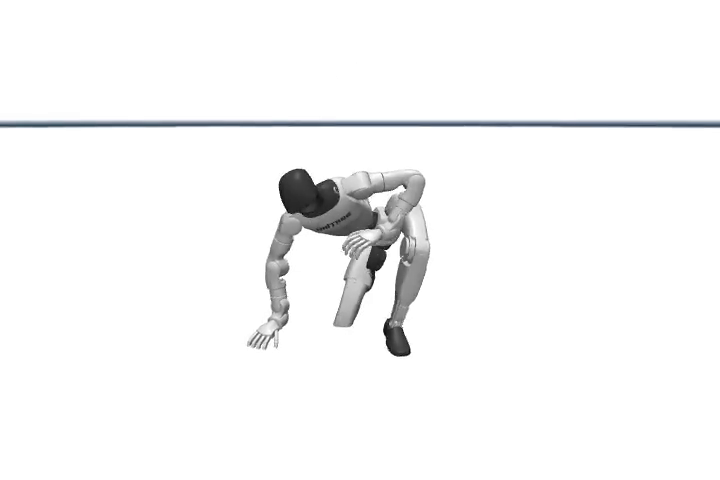}
        \caption{Getup}
    \end{subfigure}
    \begin{subfigure}[c]{0.32\linewidth}
        \centering
        \includegraphics[width=\linewidth]{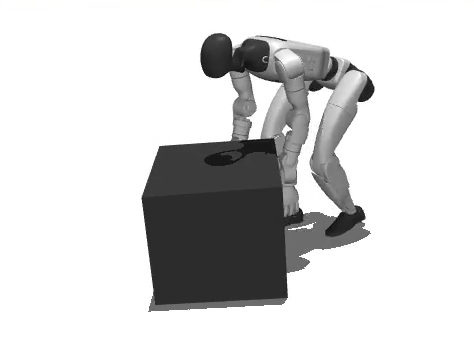}
        \caption{Pushbox}
    \end{subfigure}
\caption{Retargeting Tasks.}   
\label{fig:retarget}
\vspace{-10pt}
\end{figure}

Since MJWarp can be viewed as a GPU reimplementation of MuJoCo-CPU, it serves as a strong baseline with minimal optimizer--rollout mismatch. To evalaute ComFree-Sim, we replace MJWarp with ComFree-Sim for the optimization, while keeping  MuJoCo-CPU unchanged for the full trajectory rollout. This benchmark thus (i) characterizes the \emph{sim-to-sim gap} between ComFree and the MuJoCo contact backend in control tasks, and (ii) highlights the computational efficiency gains of ComFree-Sim when used inside the MPPI optimization loop. We retarget five Unitree G1 motions from the reference dataset (Fig.~\ref{fig:retarget}): \textbf{Cartwheels}, a dynamic acrobatic motion with continuous lateral rotation; \textbf{Martial Arts}, kicks/punches and stance transitions requiring balance; \textbf{Dance}, rhythmic whole-body motion with coordinated footwork and expressive arm gestures; \textbf{Getup}, recovery from fallen to standing; and \textbf{Pushbox}, sustained box pushing while maintaining locomotion stability.



We evaluate performance using multiple metrics that capture both motion accuracy and computational efficiency. In all tasks, we choose $k_{\text{user}} = 0.55, \ b_{\text{user}}=0.55 \times 10^{-3}$ and follow other parameters settings in SPIDER. Specifically, we report the robot base pose difference (position and quaternion), and joint configuration difference between the retargeted motion and the reference trajectory. For the Pushbox task, we additionally measure the object pose differences to assess contact interaction fidelity. Finally, we record the average computation time for one run of MPPI optimization to evaluate computational efficiency. The metrics of all tasks is shown in Table \ref{tab:retargeting}. ComFree-Sim achieves performance comparable to MJWarp on the humanoid tracking tasks, with minor differences in all metrics.  Notably, ComFree-Sim consistently achieves faster optimization time, demonstrating better computational efficiency, except for PushBox task. 

Since MJWarp is a GPU reimplementation of MuJoCo-CPU, it is expected to be more closely aligned with the MuJoCo contact backend and thus provide a strong baseline for task-level performance. Nevertheless, the results indicate that ComFree-Sim exhibits a \emph{manageable} sim-to-sim gap relative to this mature, optimization-based MuJoCo solver while delivering substantially lower rollout latency, suggesting that analytical contact resolution can remain competitive for control tasks without incurring the heavy computational cost of iterative contact solvers.

\section{Conclusion}
This work introduced \textbf{ComFree-Sim}, a GPU-parallelized analytical contact physics engine based on complementarity-free contact modeling. By resolving contact impulses in closed form with a dual-cone impedance update, ComFree-Sim achieves \emph{near-linear} runtime scaling with contact count while maintaining physical fidelity comparable to MJWarp. Extensive benchmarks validate its accuracy, stability, and 2--3 times higher throughput in dense contact scenes.  Real-world experiments on MPPI-based dexterous manipulation show that low-latency simulation yields higher control success rates and enables practical high-frequency deployment in contact-rich tasks. Dynamics-aware  retargeting experiments further show that  ComFree-Sim  achieves comparable task-level tracking with manageable sim-to-sim gap to  MuJoCo backend, while substantially reducing optimization time.



\section*{ACKNOWLEDGMENT}
We thank the MuJoCo Warp (MJWarp) team at Google DeepMind and NVIDIA for making the code publicly available. We also thank Vamsi Sai Abhijit Tadepalli from the IRIS Lab for maintaining the vision-tracking module used in our real-world in-hand manipulation experiments.

\bibliographystyle{IEEEtran}
\bibliography{references}

\end{document}